\documentclass[11pt,a4paper]{article}
\usepackage[hyperref]{acl2021}
\usepackage{graphicx}
\usepackage{times}
\usepackage{latexsym}
\usepackage[noend]{algpseudocode}
\usepackage{amsmath}
\usepackage{footmisc}
\usepackage{afterpage}
\usepackage{subcaption}

\usepackage{hyperref}
\hypersetup{
    colorlinks=true,
    linkcolor=blue,
    filecolor=magenta,      
    urlcolor=blue,
}

\usepackage{microtype}

\aclfinalcopy 

\title{LRG at SemEval-2021 Task 4: Improving Reading Comprehension with Abstract Words using Augmentation, Linguistic Features and Voting} 

\author{Abheesht Sharma\thanks{\scriptsize\hspace{2mm}Equal contribution. Author ordering determined by coin flip.} \\
Dept. of CS\&IS \\
BITS Pilani, Goa Campus \\
 \texttt{\scriptsize f20171014@goa.bits-pilani.ac.in}
 
\And

Harshit Pandey\footnotemark[1] \\
    Dept. of Computer Science \\
    Pune University \\
    \texttt{\scriptsize hp2pandey1@gmail.com} 

\AND

Gunjan Chhablani\footnotemark[1] \\
Dept. of CS\&IS \\
BITS Pilani, Goa Campus \\
 \texttt{\scriptsize chhablani.gunjan@gmail.com}

\And
 Yash Bhartia \\
  Dept. of CS\&IS\\
  BITS Pilani, Goa Campus\\
  \texttt{\scriptsize f20190151@goa.bits-pilani.ac.in}
  
\And
 Tirtharaj Dash \\
 APP Centre for AI Research\\
  BITS Pilani, Goa Campus\\
  \texttt{\scriptsize tirtharaj@goa.bits-pilani.ac.in}
}

\date{}

\begin{document}
\maketitle
\begin{abstract}
In this article, we present our methodologies for SemEval-2021 Task-4: Reading Comprehension of Abstract Meaning. Given a fill-in-the-blank-type question and a corresponding context, the task is to predict the most suitable word from a list of 5 options. There are three sub-tasks within this task: Imperceptibility (subtask-I), Non-Specificity (subtask-II), and Intersection (subtask-III). We use encoders of transformers-based models pre-trained on the masked language modelling (MLM) task to
build our Fill-in-the-blank (FitB) models. Moreover, to model imperceptibility, we define certain linguistic features, and to model non-specificity, we leverage information from hypernyms and hyponyms provided by a lexical database. Specifically, for non-specificity, we try out augmentation techniques, and other statistical techniques. We also propose variants, namely \emph{Chunk Voting} and \emph{Max Context}, to take care of input length restrictions for BERT, etc. Additionally, we perform a thorough ablation study, and use Integrated Gradients to explain our predictions on a few samples. Our best submissions achieve accuracies of 75.31\% and 77.84\%, on the test sets for subtask-I and subtask-II, respectively. For subtask-III, we achieve accuracies of 65.64\% and 62.27\%. The code is available \href{https://github.com/gchhablani/ReCAM.git}{here}.
\end{abstract}

\section{Introduction}
\label{sec:introduction}
A very common assessment in schools is question-answering based on a given
``comprehension passage''. Students are given a comprehension passage, from which they are supposed to glean necessary information, and answer short questions (such as fill-in-the-blanks-type question) based on what they have garnered from the given passage. While trying to find the most appropriate word for the blank, the children look at the words surrounding the blank (``context"). The word should be such that
when the word fills the blank, the sentence makes sense and it is grammatically correct. 
Inspired by this, and perhaps, after the enormous success of Transformers \cite{Vaswani:17}, researchers at Google came up with a large number of ``pretraining tasks" and built knowledge-heavy language models which could be fine-tuned on various natural language processing (NLP) downstream tasks. One of the earlier pretraining tasks was ``Masked Language Modelling (MLM)", one of the two pretraining tasks of the breakthrough model, BERT \cite{devlin-etal-2019-bert}. The approach here was similar to how kids are taught language at school: some tokens in the text were randomly ``masked" and the model was trained to predict these masked tokens.

SemEval-2021 Task-4 \cite{zheng-2021-semeval-task4} focuses on a similar idea. Every sample has an article, and a corresponding question. The question has a blank which the model is supposed to predict from a set of 5 options. The novelty in the task lies in its 3 subtasks: Imperceptibility (subtask-I), Non-Specificity (subtask-II), and Intersection (subtask-III). A description of these subtasks is given in Section \ref{sec:methodology}.
In this work, we propose using BERT and its derivative models such as DistilBERT \cite{Sanh:19}, ALBERT \cite{Lan:19} and RoBERTa \cite{Liu:19}. Further, we propose 2 BERT variants: (1) \emph{BERT Voting}; (2) \emph{BERT Max. Context}. Most importantly, we also model the concepts of imperceptibility and non-specificity. For imperceptibility, we create statistical embeddings using features that have a high correlation with concreteness. For non-specificity, we propose two approaches: (1) we augment the dataset by replacing some nouns in the article by their hypernyms; and (2) we use the options' hyponyms to decide the most appropriate option. We also experiement with GA-Reader \cite{dhingra2017gatedattention} and GSAMN-based approaches \cite{lai-etal-2019-gated} by trying out their various combinations with BERT.

In Section \ref{sec:background}, we perform a succinct literature survey. Section \ref{sec:methodology} elucidates our approach, including the modelling aspect, the various variants of the base model, and the different ways we model imperceptibility and non-specificity. In Section \ref{sec:experimental_setup}, we perform an extensive ablation and comparative study.

\section{Background}
\label{sec:background}
The advent of large-scale question answering systems began with straightforward tasks, like the one introduced by the SimpleQuestions Dataset \cite{bordes2015largescale}, which consisted of knowledge-base fact triples which were later used to answer questions. However, this dataset would only judge a model based on the ability to relate the facts to the question at hand. The purpose of NLP research is to be able to create a generalised model that may answer questions based on any context, thus datasets like the CNN Daily Mail \cite{hermann2015teaching} and SQuAD \cite{rajpurkar2016squad} were created. In a typical question-answering dataset, an original and anonymised context is provided before each question. Before transformers, methods consisting of LSTM/GRUs were used to achieve good results on the aforementioned tasks. These datasets however, always had answers in the passage. 

The CLOTH \cite{xie-etal-2018-large} dataset focuses on passages from middle-school and high-school text, with multiple fill-in-the-blanks in the passage. The ReCAM \cite{zheng-2021-semeval-task4} dataset puts a twist to archetypal fill-in-the-blank datasets by providing answer choices that are abstract in some form and which are not available in the passage itself. The models created for the QA task have to take into account semantic relations between the options and the context. GA-Reader \cite{dhingra2017gatedattention}, is one such model, which utilises a multi-hop architecture with a novel attention mechanism, that serves as a baseline to this task.

\section{Methodology}
\label{sec:methodology}
\subsection{MLM-Based Transformers for Cloze-Style QA}
\label{ssec:mlm-based_transformers}
The first model we employ follows a cloze-style question answering approach, in which we use various pretrained transformer models as encoders, followed by a decoder layer, which helps us to select the correct answer.

\begin{figure}[h]
  \centering
    \includegraphics[width=0.4\textwidth]{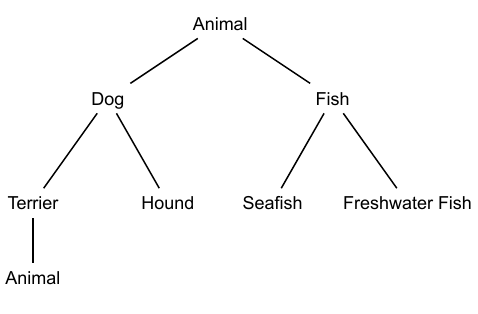}
      \caption{An example of a Hypernymy Tree}
    \label{fig:1}
\end{figure}
Specifically, we leverage BERT along with some of its popular and successful variants such as: DistilBERT, ALBERT, and RoBERTa. In the MLM task, tokens in the text are randomly masked, and the model is trained in a self-supervised way to predict these masked tokens. Conceptually, these transformers-based models are expected to take care of bidirectional context while predicting the masked token. 

\begin{figure*}
    \begin{subfigure}[t!]{0.5\linewidth}
        \centering
        \includegraphics[width=\linewidth]{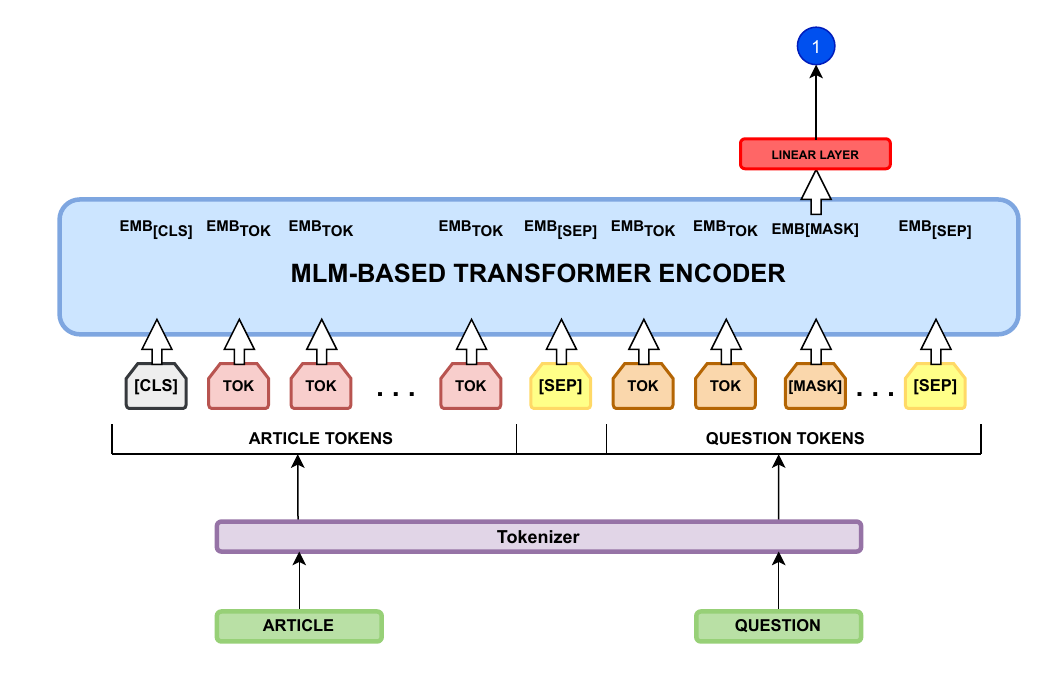}
        \caption{BERT FitB Encoder}
    \end{subfigure}
    \begin{subfigure}[t!]{0.5\linewidth}
        \centering
        \includegraphics[width=\linewidth]{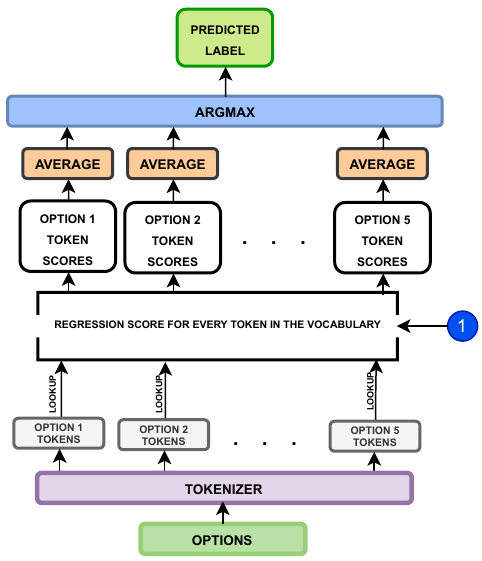}
        \caption{BERT FitB Classifier}
        
    \end{subfigure}
    \caption{Architecture of Transformer-based FitB Model}
    \label{fig:2}
\end{figure*}

In our method, firstly, the transformer model learn the contextual embeddings of the article and the question. For the next block, the embedding of the masked token (i.e., the blank) is passed through a fully-connected layer, of which, the number of outputs corresponds to the size of the vocabulary space for the pretrained model. Each candidate option is first tokenised using  WordPiece tokeniser \cite{Wu:16}, and mapped to the vector in the output vocabulary space. If the candidate option generates multiple tokens, we average the mapped scores. The model chooses the option with the highest logit value.  An overview of the model is given in Figure \ref{fig:2}.

\subsection{Improvement Approaches}
\label{ssec:improvement_approaches}
\subsubsection{Imperceptibility:}
Nouns can be clearly demarcated into two broad categories: Concrete Nouns, and Abstract Nouns. Concrete Nouns are words that represent tangible concepts, i.e., any noun referring to a name, place, object, material, etc. is considered a concrete word. Concrete words refer to concepts that can be felt by 5 human senses: Sight, Sound, Smell, Taste, and Touch.
In contrast, any noun alluding to an abstract concept that cannot be experienced by our senses is an abstract word \cite{Spreen:66}. In subtask-I, the model has to predict the most accurate and the most imperceptible word from the given options.
To model the imperceptibility of every word, we incorporate certain linguistic features which are highly correlated with the notion of ``imperceptibility". These linguistic features are listed below:

    \paragraph{Length and Frequency of the Word} In existing literature, authors have claimed that there exists strong evidence that concrete words are, in general, shorter than abstract words \cite{Tanaka:13}. A reasonable justification provided is that more frequently used words tend to be short \cite{Feng:11} 
    and since humans have a penchant for describing objects, places, or things near them, these frequently used words are generally concrete nouns. 
    It is rather intuitive that humans would prefer ease in the pronunciation of oft-used words. 
    Moreover, many abstract words in the English language are formed by adding suffixes to the root word, such as ``coarse" becomes ``coarseness", ``forget" becomes ``forgetfulness" and so on \cite{Tanaka:13}.

    \paragraph{Number of Senses of the Word} In Linguistics, polysemy refers to the capacity for a word to have multiple meanings or senses. Abstract nouns are observed to be more ``polysemous" than concrete nouns \cite{Tanaka:13}. For example, in WordNet \cite{Fellbaum:98}, the word ``dog" has 8 senses, while the word ``love" has 10 senses. 
    
    \paragraph{Number of Hyponyms} \citealt{Tanaka:13} find a direct correlation between the abstractness of a noun and the number of hyponyms the word has. We consider the number of hyponyms of the most commonly occurring sense of the word, and the average number of hyponyms of all the senses of the word.
    
    \paragraph{Score-based Features} 
    Abstract nouns evoke emotions in humans. 
    SentiWordNet \cite{baccianella-etal-2010-sentiwordnet}, another lexical database like WordNet, gives scores based on the how positive, negative or objective they are. Abstract words have a higher positive/negative score, while concrete words have a higher objective score. Again, here, we consider these scores for the most commonly occurring sense, and the average scores of all the senses of the word.
    
    \paragraph{Depth in Hypernymy Tree} This feature is more suited for non-specificity. However, we include this as a feature of imperceptibility since the concepts of imperceptibility and non-specificity are related. For example, consider the words "money" and "property". The latter is more imperceptible and non-specific than the former. Moreover, this is particularly useful for Subtask-III. Therefore, the depth of a word in the hypernymy tree is directly proportional to the concreteness of the word.

From the features above, we have a 13-dimensional vector for every word in the lexicon. The embedding is created so that every dimension is directly proportional to the concreteness of the word. For example, the length of a word is in general, indirectly proportional to the concreteness of the word, so we take the length dimension of the vector as ${large \; value - length \; of\; word}$, where we take $10,000$ as the $large \; value$. The large value chosen was the same for all features which are indirectly proportional to concreteness.

Towards improving the trained model, we use a method which we term as the \emph{Difference Method}. If the difference of the top-2 probabilities predicted by the model is greater than a certain threshold, this implies that the model is sure of the prediction it has made. However, if the difference is less than the tunable threshold, the model is ambivalent about whether the option with the highest probability or the option with the second highest probability is correct. In this case, we compute for how many dimensions the value of the linguistic embedding of the second word is less than the value of the linguistic embedding of the first word. If the majority of the values (i.e., 7) are less, we change the prediction of the model to the second-most probable option. The threshold is tuned on the dev set.
Furthermore, we use a \emph{Threshold Method}
towards improving the model performance. If the highest probability is less than a tunable threshold, the model is unsure of its predictions and we consider the improvement approaches on the option with the second-highest probability.
\subsubsection{Non-Specificity}
According to \citealp{Spreen:66}, a highly specific word refers to a very particular instance, while a non-specific word refers to a generic concept, i.e., it encompasses many classes/instances. For example, consider the words ``animal", ``bird" and ``eagle". The words are listed in increasing order of specificity.

We find parallels between the definition of specificity/non-specificity and the linguistic phenomenon of hypernymy. \citealp{Schreuder:95} define a hypernym as ``a word with a general meaning that has basically the same meaning of a more specific word". The more specific word is the corresponding hyponym.
In simpler terms, each word is related to some super-types and sub-types, called as hypernyms and hyponyms, respectively. In linguistics, hyponymy is a semantic relation between a hyponym denoting a subtype and a hypernym denoting a supertype.

For example, in figure \ref{fig:1}, as we traverse up the hypernymy tree, assuming we consider the word ``dog", we find that its hypernym is ``animal", which is much broader than ``dog". On the other hand, as we go down the hypernymy tree, we find more specific terms for the word ``dog" such as ``terrier". Essentially, hyponyms represent ``IS-A" relationships. For example, ``terrier" is a ``dog". We leverage the hypernymy property of words to help the model in deciding the most non-specific option. The two methods which we implement are:

    \paragraph{Hypernym Augmentation Method} 
    In order to infuse a sense of non-specificity (other than training on the given dataset for non-specificity), we augment the dataset for subtask-I. We randomly select $n$ nouns from the article by using a basic POS Tagging pipeline. For each noun, we use the Lesk algorithm \cite{lesk-1986-automatic} to find the most appropriate sense of the word based on its context. For this sense of the word, we find its hypernyms, pick a hypernym uniformly at random from this list of hypernyms and replace the noun in the article with the hypernym. We do this for all $2^n$ combinations, i.e., corresponding to every sample, we have $2^n$ augmented samples. Furthermore, we randomly mask tokens in this dataset and train BERT on the MLM task, on this dataset. This serves a dual purpose. Firstly, it serves as a sort of domain adaptation, and secondly, it infuses a sense of non-specificity in the model.
    
    While finetuning BERT MLM on the augmented dataset, we freeze two layers, due to time and computational constraints. We replace the normal BERT Encoder in our BERT FitB model with the BERT Encoder fine-tuned on the augmented dataset.
    
    \paragraph{Hyponyms Options Method} Here, we use the \emph{Difference Method}/\emph{Threshold Method}. If the model is sure of its prediction, we keep the prediction of the model. Otherwise, we generate hyponyms for each option using WordNet. After the hyponyms are tokenised, we use the trained model's output and map each hyponym token to the output vocabulary space and get the corresponding scores. We then take the maximum score amongst all of the hyponyms as the predicted probability for that option. The reason for incorporating this approach pertains to how the transformer models were pretrained. Consider the following sentence: ``He had a [MASK] and it was bitter". Now, suppose that we have two options: ``beer" and ``drink". Generally, our transformer-based model would look at the word ``bitter" and predict ``beer". However, ``drink" is more non-specific than ``beer".

\subsubsection{BERT Fill-in-the-blanks Variants}
To address the limitations of the vanilla transformer-based models, we attempt multiple modifications to the proposed baseline transformer models, specifically for BERT. The major limitation of the pretrained BERT model that we've used, is the restriction on the length of the tokenised inputs. Only 512 tokens from a sample can be processed by BERT in one parse and hence, some articles end up getting truncated and context is lost. The following are some of the modifications we've made to improve the performance of our models:
    \paragraph{Voting} We tokenise the question and the article. We split the article into chunks and pair each chunk with the question such that the length of the tokenised $(chunk, question)$ pair is 512. While splitting the article into chunks, we keep a max-overlap stride of 128 so that the context of the previous chunk is not lost. We train the model on these newly formed $(chunk, question)$ pairs. During inference, we take the weighted sum of the logits. For \emph{BERT FitB Voting (Similarity)}, the weights are calculated as:
\begin{equation}
    weight_{ij} = \frac{u_i.v_j}{||u_i||||v_j||}
\end{equation}
where $u_i$ is the embedding of the question in the $i^{th}$ sample, and $v_j$ is the embedding of the $j^{th}$ chunk of the sample's article. To find the embeddings, we extract the \emph{[CLS]} embedding from a pretrained BERT encoder.

We also try out an alternate way of defining the weights:
\begin{equation}
    weight_{ij} = \frac{|\{q_i\;toks.\}\; \cap \; \{chunk_j\;toks.\}|}{|\{chunk_j\;toks.\}|}
    \label{eq:weight_2}
\end{equation}
where $\{q_i\;toks.\}$ is the set of tokens in the $i^{th}$ sample's question, and $\{chunk_j\; toks.\}$ is the set of tokens in the $j^{th}$ chunk of the sample. $|.|$ represents the cardinality of a set. We call the method \emph{BERT FitB Voting (Exact Matching)}.

We normalise the computed weights:
\begin{equation}
    norm\_weight_{ij} = \frac{weight_{ij}}{\sum_{j=1}^{n_i} weight_{ij}}
\end{equation}
where $ n_i$ is the number of chunks in the $i^{th}$ sample.

The idea behind this is that higher the similarity between the question and the article's chunk, higher is the weight assigned to the logits returned by the trained model with the question-chunk pair as input. In Equation \ref{eq:weight_2}, we find the fraction of tokens common between the question and chunk.

    \paragraph{Max Context} This method is a slight modification of the Voting Method. Instead of training the model on all $(chunk, question)$ pairs for a particular sample, we train the model on the pair with the highest weight. The weights are calculated as described in Equation \ref{eq:weight_2}.
    
\begin{table*}[]
\centering
\resizebox{0.8\textwidth}{!}{%
\begin{tabular}{|c|c|c|c|c|c|}
\hline
\multicolumn{2}{|c|}{\textbf{Model Information}} & \multicolumn{2}{c|}{\textbf{Imperceptibility}} & \multicolumn{2}{c|}{\textbf{Non-Specificity}} \\ \hline
\textbf{Model}                & \textbf{Variant} & \textbf{Val Acc.}     & \textbf{Test Acc.}     & \textbf{Val Acc.}     & \textbf{Test Acc.}    \\ \hline
BERT Fill-in-the-Blank        & base             & 67.03\%               & 66.77\%                & 64.39\%               & 65.74\%               \\ \hline
BERT Fill-in-the-Blank        & large            & \textbf{74.79}\%               & \textbf{75.30}\%                & \textbf{72.73}\%               & \textbf{75.16}\%               \\ \hline
DistilBERT Fill-in-the-Blank  & base             & 67.03\%               & 66.02\%                & 63.69\%               & 62.67\%               \\ \hline
RoBERTa Fill-in-the-Blank     & base             & 52.45\%               & 51.11\%                & 33.73\%               & 35.99\%               \\ \hline
RoBERTa Fill-in-the-Blank     & large            & 51.02\%               & 52.44\%                & 33.14\%               & 34.95\%               \\ \hline
ALBERT Fill-in-the-Blank      & base-v2          & 31.42\%               & 30.46\%                & 31.84\%               & 31.14\%               \\ \hline
ALBERT Fill-in-the-Blank      & large-v2         & 31.06\%               & 30.76\%                & 30.08\%               & 33.27\%               \\ \hline
GA-Reader (baseline)           & -                & 21.23\%               & 21.51\%                & 21.50\%               & 21.86\%               \\ \hline
\end{tabular}%
}
\caption{Results of the Vanilla Fill-in-the-Blank(FitB) Models and GA-Reader}
\label{tab:vanilla-fitb-models}
\end{table*}

\subsubsection{GA-Reader-based Approaches}
We propose a few modifications to the baseline, namely GA-Reader \cite{dhingra-etal-2017-gated} provided by the organisers.

\paragraph{GA-Reader BERT}
We use GA-Reader on top of BERT embeddings. This
could lead to potential improvement in
performance for subtask-I as BERT embeddings are more feature-rich than GloVe embeddings.


\paragraph{GA-BERT}
Based on the Gated-Attention Reader, we came up with an approach that uses Gated-Attention across two-BERT streams. The first stream takes in the question input, and works like the regular BERT model. The second stream takes the article input. Assume the layer outputs for layer $L$ are $Q_{L}$ and $A_{L}$, respectively, for question and article streams. Then, to the layer $L+1$ for question stream, $Q_{L}$ is passed as input, while to layer $L+1$ for article stream, $GA(Q_{L}, A_{L})$ is passed, where $GA$ is the Gated-Attention function. This is done for all 12 layers of BERT-BASE. Finally, on this model, two types of heads are attached -  Selection and Pooling (similar to BERT FitB), and Attention Classification (similar to GA-Reader). The logits for each head are concatenated and a fully-connected layer is added on top. Since this is a major change in the architecture of BERT, this model needs a significant amount of pretraining. 


\paragraph{Answer-Attention}
Since GA-Reader also attends to the candidate answer embeddings, we also attempt an approach where we pass the options to the BERT model. On the option embeddings and the \textit{[MASK]} token embeddings, we apply multiplicative attention (dot product) to get attention scores. These scores are directly used as logits for the prediction. 

\subsubsection{GSAMN-based Approaches}
\paragraph{BERT-GSAMN-Cloze}
\citet{lai-etal-2019-gated} propose a combination of Gated-Attention and Self-Attention - Gated Self-Attention (GSA). They show improvements on smaller datasets compared to Compare-Aggregate Approaches. We use two GSA layers on top of BERT Embeddings, and use the same decoder and selection method as BERT FitB.
\afterpage{%
\begin{figure*}
  \centering
    \includegraphics[width=\textwidth]{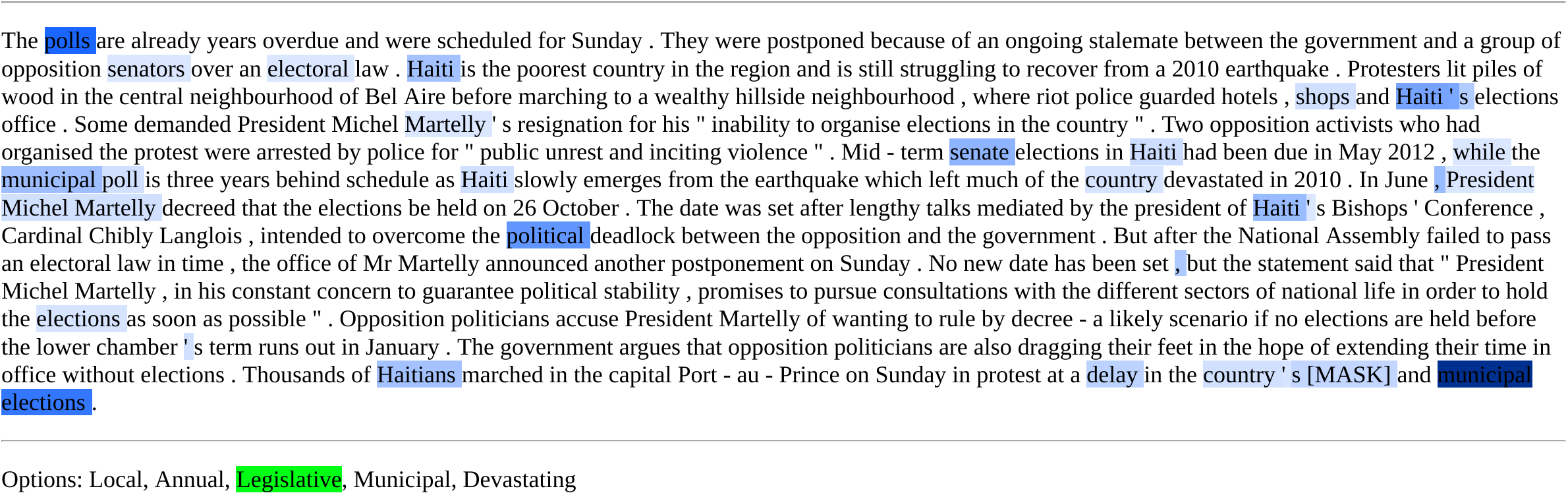}
      \caption{Explanation of a Correctly Classified Sample from Subtask-I (Imperceptibility). The correct option is highlighted in green.}
    \label{fig:3}
\end{figure*}
}

\section{Experimental Setup}
\label{sec:experimental_setup}
In all our experiments, we use the PyTorch implementations of the transformers-based models provided by the HuggingFace \cite{wolf:19}. The metric for all the 3 subtasks is accuracy. For subtask-I, to obtain the linguistic features mentioned in \ref{ssec:improvement_approaches}, and to obtain the hypernyms and hyponyms for subtask-II, we use the lexical database, WordNet provided by NLTK \cite{bird-loper-2004-nltk}, a library in Python. For both subtasks, we train our models on train + trial dataset, and evaluate them on the dev set.

The training and the evaluation of systems was on Google Colaboratory's free GPU (NVIDIA K80/P100). The training time varies with the models. It is around 1-2 hours for the base variants and 2-4 hours for the large models, which is well within the 12 hour limit of Colab. DistilBERT took about half an hour for training.

For finetuning the BERT FitB Hypr Aug Model on the augmented dataset on the MLM task, we use Nvidia-DGX Station with the following specifications: four 32 GB Tesla V100 GPUs, 256 GB RAM and forty Intel Xeon 2.20GHz processors since it is a computationally intensive task.

\subsection{Hyperparameters}
For all our experiments, we use Adam Optimiser \cite{kingma-2017-adam} and Cross Entropy Loss. For choosing the optimal set of hyperparameters, we run a Grid Search on our models. We zero in on a learning rate of 1e-5. Schedulers such as Linear Scheduler, Cosine Annealing Scheduler, etc. seem to have a negative impact on the results. For the FitB models, we keep all the layers unfrozen. Additionally, the maximum input length is kept as 512. We train our models for 4 epochs, keeping a batch size of 2.

\subsection{Ablation Study/Results}
Among the vanilla models, BERT FitB Large performs the best. This is understandable when it comes to DistilBERT and ALBERT, since these models are pruned and distilled for faster computation. Notably, DistilBERT gives comparable performance to BERT FitB Base. A slightly surprising observation was that there is a degradation in accuracy on using RoBERTa. This could be because even though it was pretrained more robustly than BERT on the MLM task, it was not pretrained on the Next Sentence Prediction Task, and hence, might perform worse on Textual Entailment tasks. A peculiar observation is that the large variants of ALBERT FitB and RoBERTa FitB models perform worse than their base variants. This may imply that more training data is needed to train the large variants. For subtask-I, in table \ref{tab:imperceptibility-improvements}, we also demonstrate the results of BERT Ensemble, in which we ensemble (i.e., averaging over the predictions) two checkpoints saved during the training process.\\
When it comes to the \emph{Difference Method} using Linguistic Features for imperceptibility, we observe an improvement on the dev set, but a slight fall is observed while evaluating it on the test set. This might be solved by careful tuning of the threshold. In the future, we aspire to learn embeddings using these Linguistic Features as input to common models such as Word2Vec \cite{mikolov-etal-2013}.

For non-specificity, with the hypernym augmentation method, BERT FitB achieves lower accuracy. A possible reason for this could be that replacing the nouns with their hypernyms in some contexts changes the meaning of the sentence (even though we use Lesk Algorithm for WSD, not all hypernyms make sense). For example, the word ``drink" is replaced with ``food". For the hyponyms method, we can improve our results by recursively generating hyponyms for a particular option, instead of taking the immediate hyponyms. Again, threshold tuning may help.
\begin{table}[]
\centering
\resizebox{\linewidth}{!}{%
\begin{tabular}{|c|c|c|c|}
\hline
\textbf{Model}     & \textbf{Variant} & \textbf{Val Acc.} & \textbf{Test Acc.} \\ \hline
BERT FitB LF       & large            & 75.75\%           & 75.06\%            \\ \hline
DistilBERT FitB LF & base             & 68.10\%           & 65.73\%            \\ \hline
BERT FitB ENS      & large            & 75.15\%           & \textbf{77.28\%}            \\ \hline
BERT FitB ENS LF   & large            & 75.87\%           & 75.26\%            \\ \hline
BERT FitB EM       & large            & 76.58\%           & 76.35\%            \\ \hline
BERT FitB EM LF    & large            & \textbf{76.82\%}           & 76.10\%            \\ \hline
BERT FitB VS       & large            & 76.58\%           & 76.54\%            \\ \hline
BERT FitB VS LF    & large            & \textbf{76.82\%}           & 76.20\%            \\ \hline
BERT MC            & large            & 74.07\%           & 73.76\%            \\ \hline
\end{tabular}%
}
\caption{Results and Ablation Study of the Improvement Methods on Subtask-I\footref{footer:1} }
\label{tab:imperceptibility-improvements}
\end{table}

In Table \ref{tab:non-spec-improvements}, a positive sign for the \emph{Difference Method} or the \emph{Threshold Method} is the improvement in the results of BERT FitB Voting (Exact Matching) when we consider the hyponyms. The accuracy jumps from 72.86\% to 75.79\% on the dev set and from 77.83\% to 78.98\% on the test set. This reinforces our claim that with more careful tuning of the threshold, we might get improvements on the test set in other methods.

\afterpage{%
\begin{figure*}
  \centering
    \includegraphics[width=\textwidth]{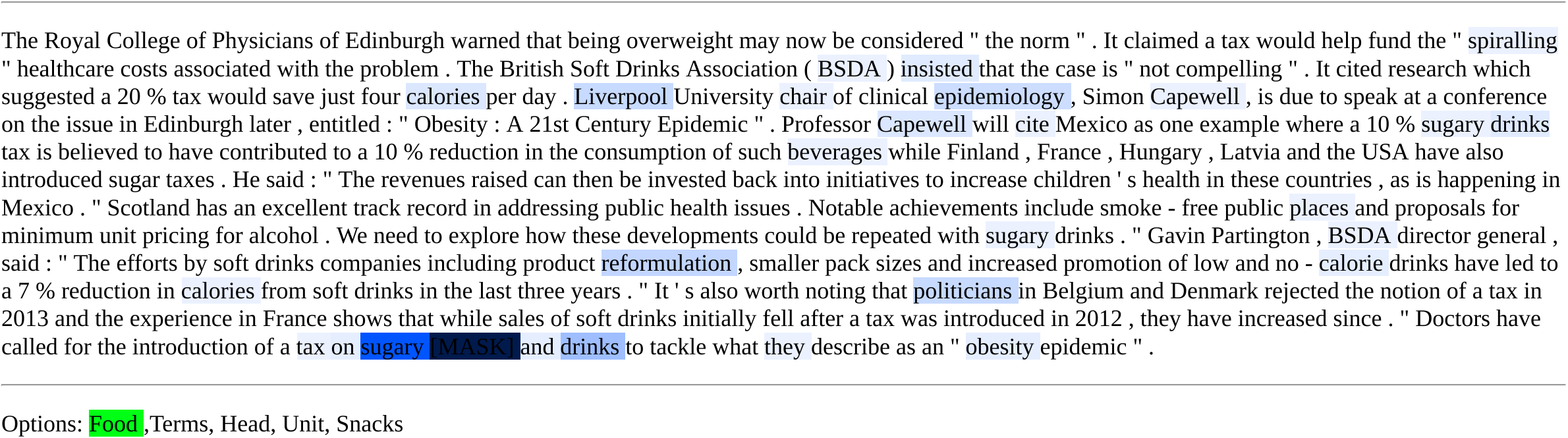}
      \caption{Explanation of a Correctly Classified Sample from Subtask-II (Non-Specificity). The correct option is highlighted in green.}
    \label{fig:4}
\end{figure*}
}

BERT FitB Voting performs better than vanilla BERT FitB on both subtasks. This is intuitive since in the latter, we truncate the article to 512 tokens without any consideration of how much context is lost. Voting, on the other hand, considers all contexts and hence, gives a superior performance.

\begin{table}[]
\centering
\resizebox{\linewidth}{!}{%
\begin{tabular}{|c|c|c|c|}
\hline
\textbf{Model}          & \textbf{Variant} & \textbf{Val Acc.} & \textbf{Test Acc.} \\ \hline
BERT FitB Hypo          & large            & 75.09\%           & 72.83\%            \\ \hline
BERT FitB Hypr Aug      & large            & 62.26\%           & 60.78\%            \\ \hline
BERT FitB Hypr Aug Hypo & large            & 64.51\%           & 55.52\%            \\ \hline
BERT FitB EM            & large            & 72.86\%           & 77.83\%            \\ \hline
BERT FitB EM Hypo       & large            & \textbf{75.79}\%           & \textbf{78.98}\%            \\ \hline
BERT FitB VS            & large            & 73.09\%           & 77.59\%            \\ \hline
BERT FitB VS Hypo       & large            & 75.56\%           & 78.63\%            \\ \hline
BERT MC                 & large            & 71.33\%           & 71.21\%            \\ \hline
\end{tabular}%
}
\caption{Results and Ablation Study of the Improvement Methods on Subtask-II\footref{footer:1}}
\label{tab:non-spec-improvements}
\end{table}

\footnotetext{LF=Linguistic Features, ENS=Ensemble, FitB=Fill-in-the-Blank, EM=Exact Matching, VS=Voting (similarity), MC=Max Context, Aug=Augmentation, Hypr=Hypernym, Hypo=Hyponym\label{footer:1}}

For GA-Reader-BERT, when compared with the GA-Reader baseline, the accuracy improves from 21\% to 39\% on subtask-I dev set. Due to computational restrictions, we couldn't pretrain GA-BERT, and only fine-tuned it for subtask-I to get an idea about its performance, which was sub-optimal (19\%). The Answer-Attention system gave us a dev score of $\approx$61\% on subtask-I, which is much higher than the baseline.

BERT-GSAMN-Cloze achieves $\approx$31\% accuracy on subtask-I dev set. The reasons for this could be lack of pretraining, unlike the original paper, or different way to getting the output logits. We see improvement as we reduced number of layers to 1($\approx$38\%) and to 0($\approx$73\%). Hence, we discarded this approach.

\begin{table*}[]
\centering
\resizebox{\textwidth}{!}{%
\begin{tabular}{|c|c|c|c|}
\hline
\multicolumn{2}{|c|}{\textbf{Imperceptibility}}               & \multicolumn{2}{c|}{\textbf{Non-Specificity}}             \\ \hline
\textbf{Model}                           & \textbf{Test Acc.} & \textbf{Model}                       & \textbf{Test Acc.} \\ \hline
BERT FitB                                & \textbf{65.64}\%            & BERT FitB                            & 61.83\%            \\ \hline
DistilBERT FitB                          & 52.16\%            & BERT FitB with Hyponyms              & 59.95\%            \\ \hline
DistilBERT FitB + Linguistic Features    & 51.61\%            & BERT FitB with Hypernym Augmentation & 45.98\%            \\ \hline
BERT FitB + Linguistic Features          & 65.54\%            & BERT FitB Voting (Exact Matching)           & \textbf{62.27}\%            \\ \hline
BERT FitB Ensemble + Linguistic Features & 64.95\%            & -                                    & -                  \\ \hline
\end{tabular}%
}
\caption{Submitted Results of Subtask-III: Testing the performance of a system that is trained on one subtask and evaluated on the other.}
\label{tab:intersection}
\end{table*}

\subsection{Analysis of BERT FitB using Integrated Gradients}
We use the method of Integrated Gradients \cite{sundararajan-17}.  We follow \citet{ramnath-etal-2020-towards} to compute the word-wise attribution scores for BERT FitB for both subtasks. We compute the Integrated Gradients of the target with respect to the embedding outputs. The Riemann Right Approximation Method with $n_{steps}=25$ is used. After obtaining the token-wise attribution scores, we obtain the word-wise attribution scores by using token-to-word offset mapping. We pick the top-10 word-wise attribution scores and normalise them. To implement IG, we use the Captum \cite{kokhlikyan-2020-captum} library. In favour of brevity, we present one example for each subtask.

In Fig. \ref{fig:3}, the correct answer is ``legislative". The attribution scores of words like \emph{senate}, \emph{senators}, \emph{municipal} and \emph{President} are high, as is demonstrated by the intensity of the colour. The word ``legislative" is, in a sense, more imperceptible than any of the words mentioned above. The \emph{senate} is the legislative branch of the government, and \emph{senators} are its members; \emph{municipal} refers to municipal corporations which are the grassroots governing bodies, etc. Moreover, other words such as \emph{elections}, \emph{political}, \emph{country} also have high attribution scores. These words are related to ``legislative" which exhibits the fact that BERT FitB is not only able to learn the concept of imperceptibility, but is also able to predict a suitable word.

Similarly, in Fig. \ref{fig:4}, the correct answer is ``food". Note that ``snacks" is also an option; however, food is more non-specific than ``snacks" and hence, food is the correct option. Another interesting thing to note is the high attribution scores for words/phrases like \emph{calories}, \emph{beverages}, \emph{sugar} and \emph{sugary drinks}. This backs the fact that the model is able to learn the concept of non-specificity, i.e., the above mentioned words are essentially hyponyms of ``food". 

\section{Conclusion}
We tried out myriad approaches, taking care to not only focus on the architecture aspect, but also how we can quantify imperceptibility and non-specificity. Although we did not achieve favourable improvements in all approaches, we did observe gains in accuracy on the dev set. We reckon that with more careful tuning of parameters such as the threshold in the \emph{Difference Method}, we will be able to achieve these gains on the test set.

We further interpreted the outputs of transformers-based models using Integrated Gradients, and demonstrated that transformer models are able to learn the concepts of imperceptibility and non-specificity. In the future, we intend to solidify our proposed approaches and carry out further research in this interesting field.
\section*{Acknowledgments}
We thank Rajaswa Patil\footnote{ \url {https://rajaswa.github.io/} } and Somesh Singh\footnote{\url {https://someshsingh22.github.io/} } for their support. We would also like to express our gratitude to our colleagues at the Language Research Group (LRG)\footnote{\url{https://lrg.saidl.in/}}, who have been with us at every stepping stone.

\bibliographystyle{acl_natbib}
\bibliography{anthology,acl2021}

\end{document}